\pdfoutput=1
\documentclass[11pt]{article}

\usepackage[preprint]{acl}

\usepackage{times}
\usepackage{latexsym}
\usepackage{todonotes}
\usepackage{booktabs}
\usepackage{placeins}
\usepackage{multirow}
\usepackage{multicol}
\usepackage{graphicx}
\usepackage{svg}
\usepackage{array}
\usepackage{tcolorbox}
\usepackage{lineno}
\usepackage[T1]{fontenc}
\usepackage[utf8]{inputenc}

\usepackage{microtype}

\usepackage{inconsolata}

\usepackage{graphicx}

\usepackage{cancel}
\definecolor{dukeblue}{RGB}{1, 33, 105}

\title{\raisebox{-0.1\height}{\includegraphics[width=0.8cm]{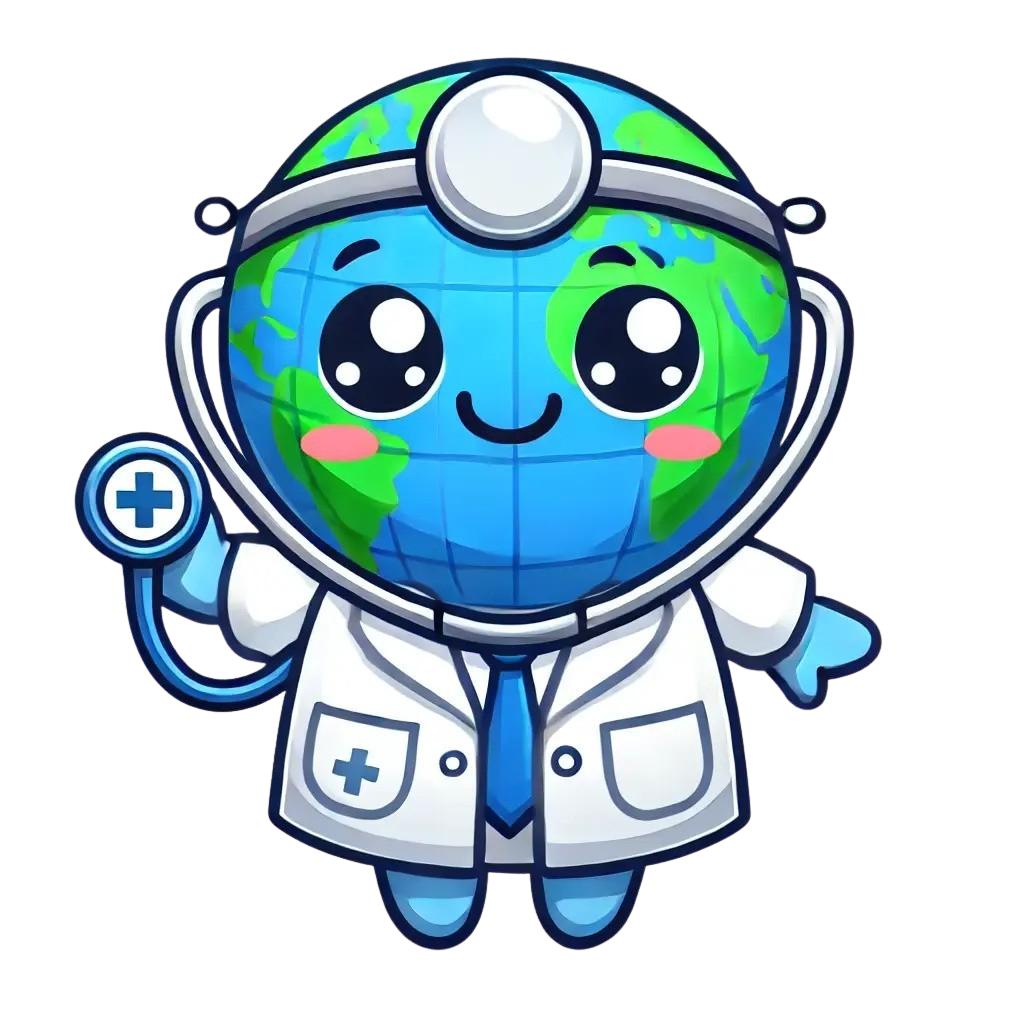}}\texttt{WorldMedQA-V}: a multilingual, multimodal medical examination\\ dataset for multimodal language models evaluation}

\author{
João Matos$^{1}$\thanks{Co-first authors: João Matos and Shan Chen}, 
Shan Chen$^{2,3,4}$\footnotemark[1], % Uses the same footnote as Shan Chen
Siena Placino$^{5}$, Yingya Li$^{2,4}$, Juan Carlos Climent Pardo$^{2,3}$ \\
\textbf{Daphna Idan$^{6}$, Takeshi Tohyama$^{7,9}$, David Restrepo$^{7}$, Luis F. Nakayama$^{7}$} \\
\textbf{Jose M. M. Pascual-Leone$^{8}$, Guergana Savova$^{2,4}$, Hugo Aerts$^{2,3,10}$, Leo A. Celi$^{2,7,11}$} \\
\textbf{A. Ian Wong$^{12}$, Danielle S. Bitterman$^{2,3,4}$, Jack Gallifant$^{2,3}$\thanks{Corresponding author: jgallifant@bwh.harvard.edu}}
\small % Reduces the font size for the entire block
\\[0.5ex] % Adds some extra space between authors and affiliations
$^1$Oxford, $^2$Harvard, $^3$Mass General Brigham, $^4$Boston Children's Hospital, \\
$^5$St. Luke's Medical Center, $^6$Ben-Gurion University of the Negev, $^7$MIT, $^8$Alcalá University, \\
$^9$International University of Health and Welfare, $^{10}$Maastricht University, $^{11}$BIDMC, $^{12}$Duke
}
\begin{document}

\maketitle

\begin{abstract}
Multimodal/vision language models (VLMs) are increasingly being deployed in healthcare settings worldwide, necessitating robust benchmarks to ensure their safety, efficacy, and fairness. Multiple-choice question and answer (QA) datasets derived from national medical examinations have long served as valuable evaluation tools, but existing datasets are largely text-only and available in a limited subset of languages and countries. To address these challenges, we present \texttt{WorldMedQA-V}, an updated multilingual, multimodal benchmarking dataset designed to evaluate VLMs in healthcare. \texttt{WorldMedQA-V} includes 568 labeled multiple-choice QAs paired with 568 medical images from four countries (Brazil, Israel, Japan, and Spain), covering original languages and validated English translations by native clinicians, respectively. Baseline performance for common open- and closed-source models are provided in the local language and English translations, and with and without images provided to the model. The \texttt{WorldMedQA-V} benchmark aims to better match AI systems to the diverse healthcare environments in which they are deployed, fostering more equitable, effective, and representative applications.\footnote{All code is accessible on \url{https://github.com/WorldMedQA/V} and  the dataset on \url{https://huggingface.com/datasets/WorldMedQA/V}.}
\end{abstract}

% Introduction @Joao
\section{Introduction}
\label{sec:introduction}

Generative artificial intelligence (AI) models are increasingly being adopted in healthcare, highlighting the need for robust benchmarks to assess their safety, efficacy, and fairness \cite{thirunavukarasu_large_2023, clusmann_future_2023, abbasian_foundation_2024, wiggers_hugging_2024}. 

One of the key evaluation tasks in Natural Language Processing (NLP) is Question Answering (QA)\cite{yu2024large,fan2023bibliometric}, which involves building systems that can automatically respond to human queries in natural language by combining language understanding with information retrieval \cite{medqa}. Multi-choice QA benchmarks have become essential not only for evaluating large language models (LLMs) but also for assessing vis language models (VLMs) in medicine \cite{Liu2024}.

\begin{figure}
    \centering
    \includegraphics[width=.95\linewidth]{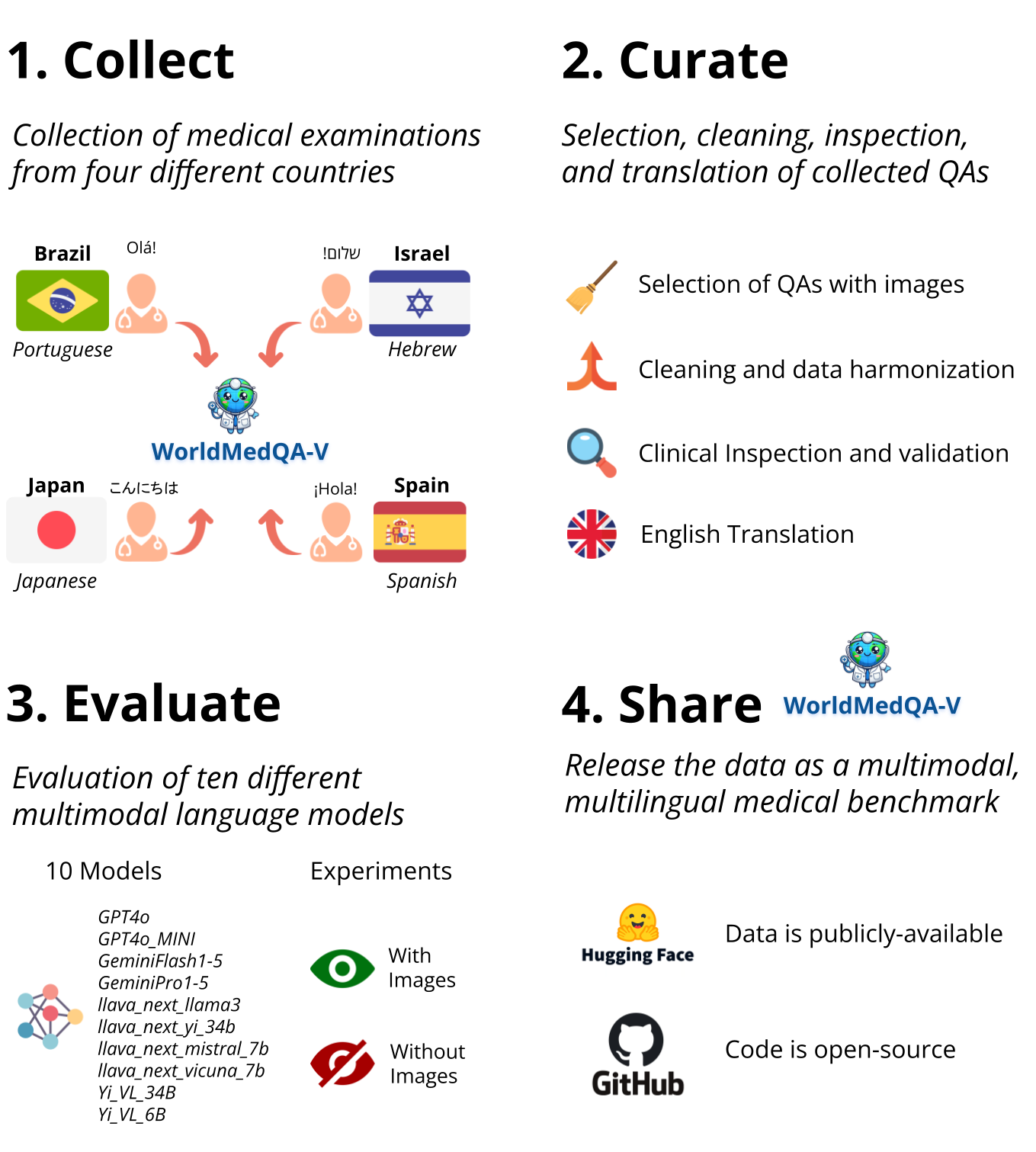}
    \caption{\texttt{WorldMedQA-V} dataset generation and evaluation workflows.}
    \label{fig:workflow}
\end{figure}

Recent research has explored the performance of LLMs in medical exams, with ChatGPT being the first AI system to pass the USMLE \cite{Kung2023}, prompting further studies \cite{gobira_performance_2023, Liu2024,Chen_2024}. A recent review identified 45 studies on ChatGPT's performance in medical exams \cite{Liu2024}, but VLMs remain underexplored in medical tasks \cite{yan2023multimodalchatgptmedicalapplications, wu2023gpt4visionservemedicalapplications}. Despite progress, current models face limitations such as context fragility, biases, and inconsistent multilingual performance \cite{gallifant_rabbits, zack_assessing_2024, chen2024crosscare}. There is also a need for more diverse datasets to ensure equitable AI evaluation in healthcare \cite{restrepo_analyzing_2024}. Key gaps include:
\begin{itemize}
    \item \textbf{Real-world validity}: Studies reveal errors in existing medical QA datasets \cite{saab2024capabilitiesgeminimodelsmedicine}.
    \item \textbf{Linguistic diversity}: Many datasets lack language representation (Appendix Table \ref{tab:medqa_datasets}) \cite{multiligual_restrepo, restrepo_analyzing_2024,ryan_unintended_2024}.
    \item \textbf{Imaging data}: Most medical QA benchmarks exclude multimodal data (Appendix Table \ref{tab:medqa_datasets})
    \item \textbf{Training data contamination}: Outdated datasets may overlap with LLM/VLM training corpora \cite{zhang_language_2024, zhang_careful_2024,gallifant_rabbits}.
\end{itemize}

To address these issues, we introduce \texttt{WorldMedQA-V}, a multilingual, multimodal dataset for evaluating language and vision models. Key contributions include:
\begin{itemize}
    \item \textbf{Multimodal medical exams from four countries}, supporting local languages and English.
    \item Previously \textbf{unseen} multimodal exam questions with \textbf{clinical validation} by medical professionals.
    \item Baseline \textbf{performance reporting of current state-of-the-art VLMs} across languages, including an evaluation of performance differentials between \textbf{local languages and English}.
    \item An investigation into the \textbf{impact of adding image data} to model performance and \textbf{stability across language} translations.
\end{itemize}

% Related Work @Joao Shan Jack Siena
\section{Related Work}
\label{sec:related}
Recent benchmarks like MMMU \citep{zhang2023mmmu}, EXAMS-V \citep{zhang2023examsv}, and CulturalVQA \citep{wang2023culturalvqa} evaluate VLMs across multiple languages and disciplines, revealing notable performance gaps across linguistic and cultural contexts. Studies show that VLMs perform better in English, likely due to the predominance of English training data \citep{adam2023generative,weidinger2021ethical}. These findings highlight the need for improving VLMs in diverse languages and cultural settings, especially in specialized domains.

\begin{figure*}
    \centering
    \includegraphics[width=.95\linewidth]{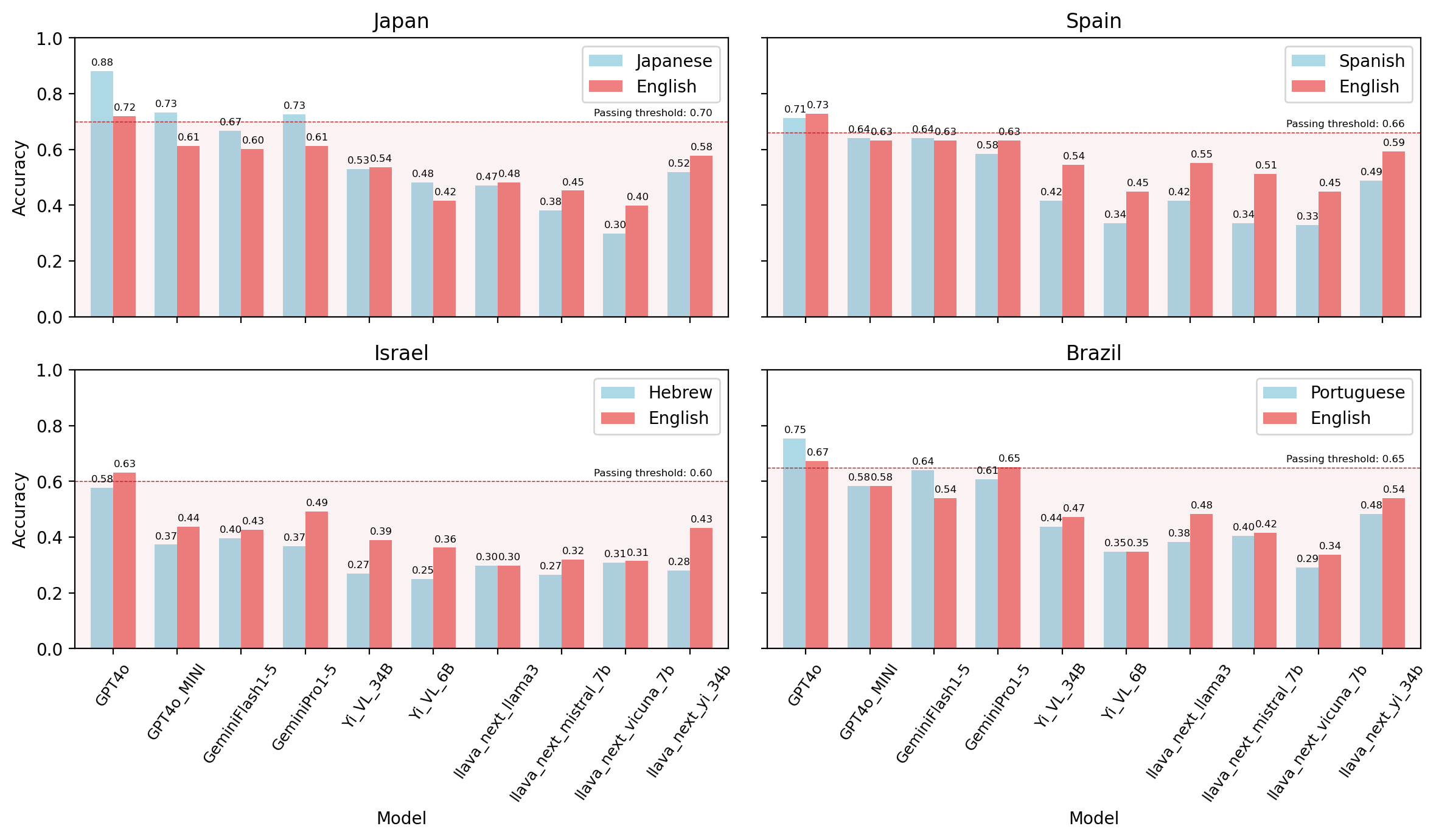}
    \caption{Accuracy in local language and English across models and countries. The red-shaded area highlights each country's exam passing threshold. Passing score is a proxy here since our dataset is a subset. Detailed results in Appendix \ref{sec:detail_performance}}
    \label{fig:model_comparison}
\end{figure*}

Appendix \ref{sec:medqas} Table \ref{tab:medqa_datasets} summarizes existing medical QA datasets by country. Six languages are covered, spanning seven countries across three continents: Asia (China, India, South Korea, and Taiwan), Europe (Spain and Sweden), and North America (U.S.).

Medical datasets from these regions highlight challenges in LLMs' performance in healthcare. In Asia, notable datasets include those from China \cite{li-etal-2021-mlec}, Taiwan \cite{medqa}, South Korea \cite{KorMedMCQA}, and India \cite{medmcqa}. The MLEC-QA dataset from China, with 136,236 multiple-choice questions, is the largest. Despite LLMs being pre-trained on vast datasets,  performance in this domain is hindered by limited diversity and quality of training data, especially for two-step reasoning and biomedical concepts \cite{li-etal-2021-mlec}. Similar trends are observed in Taiwan and South Korea, where English-pretrained models underperform on local medical exams. In Europe, datasets from Spain, Sweden, and Poland (the latter not publicly available) underscore the difficulties LLMs face, especially as question complexity increases \cite{vilares-gomez-rodriguez-2019-head}. However, recent advancements saw models like \textit{GPT3.5-Turbo} and \textit{GPT4} pass the Swedish medical licensing exam \cite{hertzberg-lokrantz-2024-medqa}, while \textit{GPT4-Turbo} slightly outperformed humans in Poland \cite{bean2024exploringlandscapelargelanguage}.

% Methods @Joao Shan Jack
\section{Methodology}
\label{sec}

Figure \ref{fig:workflow} shows the overall workflow of the study.

\subsection{Data Collection}
\label{subsec}
Our study uses medical exam data from Brazil, Israel, Japan, and Spain, consisting of multiple-choice questions from national licensing or specialization exams. Brazil’s dataset includes 100 questions per exam from the 2011–16 and 2020–24 "Revalida" exams. Israel's dataset contains 150 questions from Phase A of the resident certification exam (2020–23). Japan's data comes from the 116th–118th National Medical Licensing Examinations (2022–24), while Spain's dataset includes questions from specialization exams (2019–23). Further details are provided in Appendix \ref{sec:data_details}.
% \textbf{Brazil:} The questions were sourced from the 2011–2016 and 2020–2024 "Revalida" exams, which certify international medical graduates for practice in Brazil. Each exam has 100 QAs. Only multiple-choice QAs, covering Internal Medicine, Surgery, Pediatrics, Preventive Medicine, and Gynecology/Obstetrics, were included.

% \textbf{Israel:} This set includes questions from Internal Medicine, Clinical Microbiology, Neurology, Oncology, Ophthalmology, Urology, and Public Health, sourced from Phase A of the resident certification exam. This annual exam includes 150 questions. Data from 2020–2023 exams were used.

% \textbf{Japan:} These questions were sourced from the 116–118\textsuperscript{th} National Medical Licensing Examinations (2022–2024), publicly available on Japan’s Ministry of Health, Labour and Welfare website. These assess a wide range of medical knowledge.

% \textbf{Spain:} These questions were gathered from medical specialization exams administered by the \textit{Ministerio de Sanidad}. We used multiple-choice questions from 2019–2023.

% for camera-ready / preprint
% \subsection{Clinical Validation} A clinical validation process was carried out for all collected and translated data to ensure their quality and relevance. Native-speaking co-authors clinicians (Brazil: LFN; Israel: DI; Japan: TT; Spain: JMMPL), validated the three key stages of the process — data extraction, translation, and final question review.

\subsection{Clinical Validation} A clinical validation process was carried out for all collected and translated data to ensure their quality and relevance. Native-speaking clinicians from each country validated the three key stages of the process — data extraction, translation, and final QA review. 

\subsection{Evaluation}
% We excluded questions that allowed more than one answer or had more than one correct option, resulting in the numbers depicted in the last column of Appendix \ref{sec:data_stats}'s Table \ref{tab:data}.

% Danielle: %this is confusing. Were some questions manually adjusted to be only singe-choice MCQA and/or to include only 1 image, or were these questions thrown out? please clarify here in and the 2.1 description. I wouldn't refer to it as pre-processing since it is more than just preprocessing the text and image data if I am understadning correctly. I would refer to it as dataset curation/cleaning.

% João: % am removing this to avoid confusion. all we did was exclude the multiple image / correct answers to avoid model misbehavior. no adjustments were made to the QAs. this is mentioned in the results as our final N

\textbf{Models:} We included open- and closed-source models across a range of sizes: \textit{GPT4o-2024-05-13}, \textit{GPT4o-MINI-2024-07-18}, \textit{GeminiFlash1-5 May}, \textit{GeminiPro1-5 May}, \textit{llava-next-llama3(8B)}, \textit{llava-next-yi-34b}, \textit{llava-next-mistral-7b}, \textit{llava-next-vicuna-7b}, \textit{Yi-VL-34B}, and \textit{Yi-VL-6B}. All models were set to generate 512 tokens, with a temperature of 0 for reproducibility, and evaluated with Nvidia-GPU with CUDA > 12.0.

%Shan: Yes and the response can be covered under 512 tokens

\textbf{Experiments:} The \textit{VLMEvalKit} evaluation framework \cite{duan2024vlmevalkit} was utilized to conduct experiments. We evaluated the ten models with and without image input, using accuracy as the metric. Cohen's kappa coefficients \cite{cohen_weighted_1968} were computed to assess each model's reliability when answering the question in the original language versus the English translation.

% Results
%% General results, comparing English to local, also what Venn diagram on what is wrong @ joao
\section{Results and Discussion}

\begin{figure}[ht]
    \centering
    \includegraphics[width=.95\linewidth]{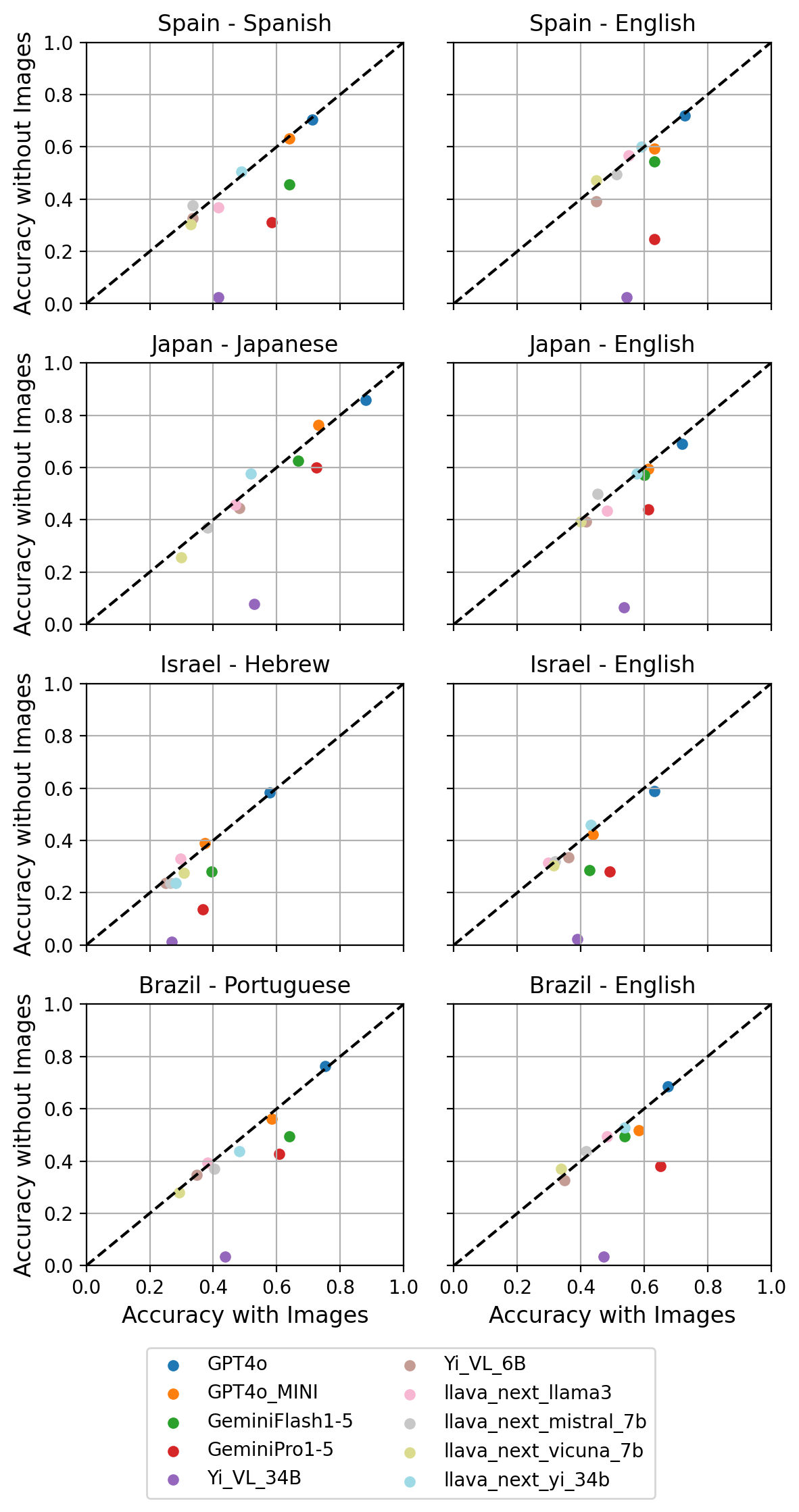}
\caption{Accuracy across countries and languages with and without image input. See Appendix \ref{sec:detail_performance} for details.}
    \label{fig:image_blind_comparison}
\end{figure}

\textbf{Dataset:} The complete \texttt{WorldMedQA-V} includes a total of 726 QAs and 850 images across four countries: Brazil, Israel, Japan, and Spain. Each QA is paired with at least one image, though some images appear in more than one question. After the exclusion of questions with multiple images or correct options, the final evaluation subset contains 568 QAs, each with a single associated image and correct option. Table \ref{tab:data}, in Appendix \ref{sec:data_stats}, provides a detailed summary of data distribution across countries and languages. Box 1 in Appendix \ref{sec:example} shows an example from the Brazilian dataset.

\textbf{VLMs' Performance:} Figure \ref{fig:model_comparison} shows model performance across datasets in the local language and in English. Compared to the previously reported performance of \textit{GPT4} on the USMLE, which is ~90\% \cite{brin_comparing_2023}, all models exhibit reduced performance when confronted with both image and text data. \textit{GPT4o} emerged as the best-performing model. The only dataset for which \textit{GPT4o} did not achieve a passing grade was the Israel dataset in Hebrew on which it achieved only 58\%. Interestingly, \textit{GPT4o} passed the English-translated version (63\%) of the Israeli dataset. The other dataset in \texttt{WorldMedQA-V} with a non-Roman alphabet is the Japan dataset, on which \textit{GPT4o} achieved an accuracy of 88\%, exceeding the 70\% passing threshold. This may be because Japanese is better represented in pretraining datasets and has character overlap with Mandarin. The underperformance in Hebrew, in contrast, could reflect Hebrew’s lower representation in pretraining data, affecting the models' ability to understand the native language as effectively \cite{ustun2024ayamodelinstructionfinetuned}. Models generally performed better on English-translated datasets, particularly for the Spain and Israel datasets. Moreover, the English-translated Israel dataset exhibits a somewhat lower overall performance when compared to other countries, which may indicate data variations that go beyond language. In the Brazilian subset, \textit{GPT4o} scored 75\% in Portuguese and 67\% in English. Similarly, in the Japanese dataset, models such as \textit{GPT4o}, \textit{GPT4o-MINI}, \textit{GeminiFlash1-5}, and \textit{GeminiPro1-5} performed better in Japanese than in English, indicating strong language support for Japanese. The lowest accuracies were from the \textit{llava-next} series, particularly on the Israel dataset, where several variants achieved nearly random accuracies in both Hebrew and English, ranging 24-46\%. (Figure \ref{fig:model_comparison})

\textbf{Accuracy with and without image input:} Models performed better with image input. This trend was consistent across most datasets, particularly for models with lower baseline performance. However, the accuracy of the \textit{GPT} models showed only minor variations — typically within 1-3\% — regardless of whether the image was provided. Models from the \textit{Gemini} family tended to be most sensitive to the exclusion of images, with improvements ranging from 4-27\% when images were provided. The \textit{Yi-VL} and \textit{llava-next} models, which generally underperformed across the board, exhibited more stochastic variations in either direction depending on image input. Lastly, it is worth noting that the \textit{Yi-VL-34b} model had almost no predictive power without images. (Figure \ref{fig:image_blind_comparison})

\textbf{Model consistency comparing English and local languages:} Table \ref{tab:agreement_images} in Appendix \ref{sec:kappas} compares model outputs in original languages to English translations using Cohen’s kappa. \textit{GPT4o} consistently achieved the highest agreement, particularly in the Brazil, Japan, and Spain datasets, with better performance in image-based settings. The highest kappa (84\%) was observed in Spain’s text-only setting, likely due to the model’s high overall accuracy. Models like \textit{GPT4o-MINI} and \textit{GeminiFlash1-5} performed well in Brazil and Spain but lagged behind \textit{GPT4o}. In contrast, \textit{Yi-VL} showed lower agreement across countries, suggesting worse cross-language consistency. Notably, \textit{GeminiPro1-5} showed an improvement in kappa, from 16.3\% to 69.3\%, when images were included in the Spanish set, demonstrating a substantial stabilizing effect of multimodal input. Overall, model cross-linguistic consistency improved with image data input.

\label{sec:kappas}

% Conclusion
\section{Conclusion}

In this work, we introduce \texttt{WorldMedQA-V}, a clinically validated, multilingual, and multimodal dataset containing medical QAs and images from Brazil, Israel, Japan, and Spain. We evaluated the performance of 10 VLMs across both local languages and English translations, highlighting performance disparities between languages and the role of multimodal data in enhancing model accuracy. Our results show that while most models performed better with image input, language diversity remains a critical challenge, particularly for underrepresented languages like Hebrew. \texttt{WorldMedQA-V} addresses key gaps in language diversity and multimodal evaluation in healthcare, providing a robust benchmark for future VLM development.

% Limitations (doesnt count toward the 4pg limit)
\newpage
\section{Limitations}

While \texttt{WorldMedQA-V} represents a significant step toward creating a multilingual, multimodal benchmark for evaluating VLMs in healthcare, several limitations must be acknowledged. 

First, the dataset, while carefully curated by trained physicians to ensure the validity of both questions and answers, remains relatively small. As we evaluated 568 multiple-choice questions and images, the sample size is limited in comparison to larger text-based benchmarks.

Second, the dataset only includes data from four countries: Brazil, Israel, Japan, and Spain, spanning three continents. This geographic limitation results in an underrepresentation of certain regions, particularly Africa, North and Central America, Oceania, and other parts of Asia.

Furthermore, although the benchmark introduces multimodal elements, it pairs only one image per question. Real-world clinical scenarios often involve multiple images from different time points or modalities, such as a sequence of X-rays, CT scans, and pathology slides. Another limitation is that text that is within images were not translated or adapted. English translations, although validated by native-speaking clinicians from each country, require further cross-validations, as these are typically nontrivial tasks.

Additionally, the lack of open-source multimodal medical language models restricts our ability to comprehensively evaluate and compare state-of-the-art health AI using \texttt{WorldMedQA-V}. Furthermore, since the models we tested were not originally trained for the medical domain, some LLMs (e.g., Gemini) refused to respond when no image was provided for certain questions, resulting in lower scores. When evaluating model performance against a passing threshold, a limitation is that our analysis relies on a limited set of multiple-choice questions with images, which may not provide consistent difficulty levels across different questions within the same exam.

Lastly, we set the underlying assumption that each question had only one correct answer, excluding cases where multiple correct answers were possible. This decision was made to simplify evaluation, but it may not reflect the inherent ambiguity and complexity found in both medical examinations and real-world medical scenarios where multiple treatment options or diagnoses can be valid.

% Acknowledgements
\section*{Acknowledgments}
\textbf{Contributions:}
The authors contributed to this work as follows: JM, SC, and JG were responsible for the manuscript writing. JM, SC, JG, SP, JCCP, DI, DB, and AIW were involved in editing. The methods design was carried out by JM, SC, YL, AIW, DB, and JG . Data collection was conducted by JM, SP, DI, LFN, and DR. Data curation was done by JM, SC, SP, JCCP, DI, TT, DR, LFN, JMMPL, and JG. SC and YL performed the LVMs experiments. Visualizations were created by JM, SC, and DB. Data interpretation was a collective effort by all authors. Clinical validation was done by LFN (Brazil), DI (Israel), TT (Japan), and JMMPL (Spain). English translation validation was handled by LFN, DI, TT, JMMPL, JCCP, JM, and JG. Funding for the research was provided by GS, HA, LAC, AIW, and DB. Study validation was under DR, LFN, GS, HA, LAC, AIW, DB, and JG.

\textbf{Funding:}
The authors acknowledge financial support from the Google PhD Fellowship (SC), the Woods Foundation (DB, SC), the National Institutes of Health (NIH) R01CA294033 (DB, SC, JG), U54CA274516-01A1 (DB, SC, JG), R01 EB017205 (LAC), U54 TW012043-01 (DS-I Africa; LAC), OT2OD032701 (Bridge2AI; LAC), U54MD012530 (REACH Equity; AIW), the National Science Foundation (NSF) ITEST 2148451 (LAC), the ASTRO-ACS Clinician Scientist Development Grant ASTRO-CSDG-24-1244514 (DB), and a Clarendon Scholarship from the University of Oxford (JM). The authors also thank Google Cloud for funding Gemini API inference costs.

\newpage
% Bibliography entries for custom only
\bibliography{custom}

\begin{thebibliography}{47}
\providecommand{\natexlab}[1]{#1}

\bibitem[{Abbasian et~al.(2024)Abbasian, Khatibi, Azimi, Oniani, Shakeri Hossein~Abad, Thieme, Sriram, Yang, Wang, Lin, Gevaert, Li, Jain, and Rahmani}]{abbasian_foundation_2024}
Mahyar Abbasian, Elahe Khatibi, Iman Azimi, David Oniani, Zahra Shakeri Hossein~Abad, Alexander Thieme, Ram Sriram, Zhongqi Yang, Yanshan Wang, Bryant Lin, Olivier Gevaert, Li-Jia Li, Ramesh Jain, and Amir~M. Rahmani. 2024.
\newblock \href {https://doi.org/10.1038/s41746-024-01074-z} {Foundation metrics for evaluating effectiveness of healthcare conversations powered by generative {AI}}.
\newblock \emph{npj Digital Medicine}, 7(1):1--14.

\bibitem[{Adam et~al.(2023)}]{adam2023generative}
Dillon~C Adam et~al. 2023.
\newblock Generative ai for infectious diseases: An evaluation of chatgpt for medical translation.
\newblock \emph{PLOS Global Public Health}, 3(6):e0001673.

\bibitem[{{Association}()}]{ima_english}
{Israel}~{Medicine} {Association}.
\newblock \href {https://www.ima.org.il/ENG/Default.aspx} {{IMA} - {Israel} {Medicine} {Association}}.

\bibitem[{Association()}]{hebrew_data}
The Israeli~Medical Association.
\newblock \href {https://www.ima.org.il/InternesNew/ViewCategory.aspx?CategoryId=14171} {The interns' website | written examination questionnaires - stage a}.

\bibitem[{Bean et~al.(2024)Bean, Korgul, Krones, McCraith, and Mahdi}]{bean2024exploringlandscapelargelanguage}
Andrew~M. Bean, Karolina Korgul, Felix Krones, Robert McCraith, and Adam Mahdi. 2024.
\newblock \href {https://arxiv.org/abs/2310.07225} {Exploring the landscape of large language models in medical question answering}.
\newblock \emph{Preprint}, arXiv:2310.07225.

\bibitem[{Brin et~al.(2023)Brin, Sorin, Vaid, Soroush, Glicksberg, Charney, Nadkarni, and Klang}]{brin_comparing_2023}
Dana Brin, Vera Sorin, Akhil Vaid, Ali Soroush, Benjamin~S. Glicksberg, Alexander~W. Charney, Girish Nadkarni, and Eyal Klang. 2023.
\newblock \href {https://doi.org/10.1038/s41598-023-43436-9} {Comparing {ChatGPT} and {GPT}-4 performance in {USMLE} soft skill assessments}.
\newblock \emph{Scientific Reports}, 13:16492.

\bibitem[{Chen et~al.(2024{\natexlab{a}})Chen, Gallifant, Gao, Moreira, Munch, Muthukkumar, Rajan, Kolluri, Fiske, Hastings, Aerts, Anthony, Celi, Cava, and Bitterman}]{chen2024crosscare}
Shan Chen, Jack Gallifant, Mingye Gao, Pedro Moreira, Nikolaj Munch, Ajay Muthukkumar, Arvind Rajan, Jaya Kolluri, Amelia Fiske, Janna Hastings, Hugo Aerts, Brian Anthony, Leo~Anthony Celi, William G.~La Cava, and Danielle~S. Bitterman. 2024{\natexlab{a}}.
\newblock \href {https://arxiv.org/abs/2405.05506} {Cross-care: Assessing the healthcare implications of pre-training data on language model bias}.
\newblock \emph{Preprint}, arXiv:2405.05506.

\bibitem[{Chen et~al.(2024{\natexlab{b}})Chen, Li, Lu, Van, Aerts, Savova, and Bitterman}]{Chen_2024}
Shan Chen, Yingya Li, Sheng Lu, Hoang Van, Hugo J W~L Aerts, Guergana~K Savova, and Danielle~S Bitterman. 2024{\natexlab{b}}.
\newblock \href {https://doi.org/10.1093/jamia/ocad256} {Evaluating the chatgpt family of models for biomedical reasoning and classification}.
\newblock \emph{Journal of the American Medical Informatics Association}, 31(4):940–948.

\bibitem[{Clusmann et~al.(2023)Clusmann, Kolbinger, Muti, Carrero, Eckardt, Laleh, Löffler, Schwarzkopf, Unger, Veldhuizen, Wagner, and Kather}]{clusmann_future_2023}
Jan Clusmann, Fiona~R. Kolbinger, Hannah~Sophie Muti, Zunamys~I. Carrero, Jan-Niklas Eckardt, Narmin~Ghaffari Laleh, Chiara Maria~Lavinia Löffler, Sophie-Caroline Schwarzkopf, Michaela Unger, Gregory~P. Veldhuizen, Sophia~J. Wagner, and Jakob~Nikolas Kather. 2023.
\newblock \href {https://doi.org/10.1038/s43856-023-00370-1} {The future landscape of large language models in medicine}.
\newblock \emph{Communications Medicine}, 3(1):1--8.

\bibitem[{Cohen(1968)}]{cohen_weighted_1968}
Jacob Cohen. 1968.
\newblock \href {https://doi.org/10.1037/h0026256} {Weighted kappa: {Nominal} scale agreement provision for scaled disagreement or partial credit}.
\newblock \emph{Psychological Bulletin}, 70(4):213--220.

\bibitem[{de~Estudos~e Pesquisas Educacionais Anísio Teixeira {\textbar}~Inep()}]{brazil_data}
Instituto~Nacional de~Estudos~e Pesquisas Educacionais Anísio Teixeira {\textbar}~Inep.
\newblock \href {https://www.gov.br/inep/pt-br/acesso-a-informacao/perguntas-frequentes/exame-nacional-de-revalidacao-de-diplomas-medicos-expedidos-por-instituicoes-de-educacao-superior-estrangeira-revalida} {Exame {Nacional} de {Revalidação} de {Diplomas} {Médicos} {Expedidos} por {Instituições} de {Educação} {Superior} {Estrangeira} ({Revalida})}.

\bibitem[{de~Sanidad()}]{spain_data}
Ministerio de~Sanidad.
\newblock \href {https://fse.mscbs.gob.es/fseweb/view/public/datosanteriores/cuadernosExamen/busquedaConvocatoria.xhtml} {Formación {Sanitaria} {Especializada}}.

\bibitem[{Duan et~al.(2024)Duan, Yang, Qiao, Fang, Chen, Liu, Dong, Zang, Zhang, Wang, Lin, and Chen}]{duan2024vlmevalkit}
Haodong Duan, Junming Yang, Yuxuan Qiao, Xinyu Fang, Lin Chen, Yuan Liu, Xiaoyi Dong, Yuhang Zang, Pan Zhang, Jiaqi Wang, Dahua Lin, and Kai Chen. 2024.
\newblock \href {https://arxiv.org/abs/2407.11691} {Vlmevalkit: An open-source toolkit for evaluating large multi-modality models}.
\newblock \emph{Preprint}, arXiv:2407.11691.

\bibitem[{Fan et~al.(2023)Fan, Li, Ma, Lee, Yu, and Hemphill}]{fan2023bibliometric}
Lizhou Fan, Lingyao Li, Zihui Ma, Sanggyu Lee, Huizi Yu, and Libby Hemphill. 2023.
\newblock \href {https://arxiv.org/abs/2304.02020} {A bibliometric review of large language models research from 2017 to 2023}.
\newblock \emph{Preprint}, arXiv:2304.02020.

\bibitem[{Gallifant et~al.(2024)Gallifant, Chen, Moreira, Munch, Gao, Pond, Celi, Aerts, Hartvigsen, and Bitterman}]{gallifant_rabbits}
Jack Gallifant, Shan Chen, Pedro Moreira, Nikolaj Munch, Mingye Gao, Jackson Pond, Leo~Anthony Celi, Hugo Aerts, Thomas Hartvigsen, and Danielle Bitterman. 2024.
\newblock \href {https://arxiv.org/abs/2406.12066} {Language models are surprisingly fragile to drug names in biomedical benchmarks}.
\newblock \emph{Preprint}, arXiv:2406.12066.

\bibitem[{Gobira et~al.(2023)Gobira, Nakayama, Moreira, Andrade, Regatieri, and Belfort~Jr.}]{gobira_performance_2023}
Mauro Gobira, Luis~Filipe Nakayama, Rodrigo Moreira, Eric Andrade, Caio Vinicius~Saito Regatieri, and Rubens Belfort~Jr. 2023.
\newblock \href {https://doi.org/10.1590/1806-9282.20230848} {Performance of {ChatGPT}-4 in answering questions from the {Brazilian} {National} {Examination} for {Medical} {Degree} {Revalidation}}.
\newblock \emph{Revista da Associação Médica Brasileira}, 69(10):e20230848.

\bibitem[{Hertzberg and Lokrantz(2024{\natexlab{a}})}]{hertzberg-lokrantz-2024-medqa}
Niclas Hertzberg and Anna Lokrantz. 2024{\natexlab{a}}.
\newblock \href {https://aclanthology.org/2024.lrec-main.975} {{M}ed{QA}-{SWE} - a clinical question {\&} answer dataset for {S}wedish}.
\newblock In \emph{Proceedings of the 2024 Joint International Conference on Computational Linguistics, Language Resources and Evaluation (LREC-COLING 2024)}, pages 11178--11186, Torino, Italia. ELRA and ICCL.

\bibitem[{Hertzberg and Lokrantz(2024{\natexlab{b}})}]{swemedqa}
Niclas Hertzberg and Anna Lokrantz. 2024{\natexlab{b}}.
\newblock \href {https://aclanthology.org/2024.lrec-main.975} {{M}ed{QA}-{SWE} - a clinical question {\&} answer dataset for {S}wedish}.
\newblock In \emph{Proceedings of the 2024 Joint International Conference on Computational Linguistics, Language Resources and Evaluation (LREC-COLING 2024)}, pages 11178--11186, Torino, Italia. ELRA and ICCL.

\bibitem[{{Japanese Ministry of Health Labour and Welfare 2022}()}]{jp116}
{Japanese Ministry of Health Labour and Welfare 2022}.
\newblock \href {https://www.mhlw.go.jp/seisakunitsuite/bunya/kenkou_iryou/iryou/topics/tp220421-01.html} {{The 116th National Medical Examination Questions and Answers}}.

\bibitem[{{Japanese Ministry of Health Labour and Welfare 2023}()}]{jp117}
{Japanese Ministry of Health Labour and Welfare 2023}.
\newblock \href {https://www.mhlw.go.jp/seisakunitsuite/bunya/kenkou_iryou/iryou/topics/tp230502-01.html} {{The 117th National Medical Examination Questions and Answers}}.

\bibitem[{{Japanese Ministry of Health Labour and Welfare 2024}()}]{jp118}
{Japanese Ministry of Health Labour and Welfare 2024}.
\newblock \href {https://www.mhlw.go.jp/seisakunitsuite/bunya/kenkou_iryou/iryou/topics/tp240424-01.html} {{The 118th National Medical Examination Questions and Answers}}.

\bibitem[{Jin et~al.(2020)Jin, Pan, Oufattole, Weng, Fang, and Szolovits}]{medqa}
Di~Jin, Eileen Pan, Nassim Oufattole, Wei{-}Hung Weng, Hanyi Fang, and Peter Szolovits. 2020.
\newblock \href {https://arxiv.org/abs/2009.13081} {What disease does this patient have? {A} large-scale open domain question answering dataset from medical exams}.
\newblock \emph{CoRR}, abs/2009.13081.

\bibitem[{Kung et~al.(2023)Kung, Cheatham, Medenilla, Sillos, De~Leon, Elepaño, Madriaga, Aggabao, Diaz-Candido, Maningo, and Tseng}]{Kung2023}
T.~H. Kung, M.~Cheatham, A.~Medenilla, C.~Sillos, L.~De~Leon, C.~Elepaño, M.~Madriaga, R.~Aggabao, G.~Diaz-Candido, J.~Maningo, and V.~Tseng. 2023.
\newblock \href {https://doi.org/10.1371/journal.pdig.0000198} {Performance of chatgpt on usmle: Potential for ai-assisted medical education using large language models}.
\newblock \emph{PLOS Digital Health}, 2(2):e0000198.

\bibitem[{Kweon et~al.(2024)Kweon, Choi, Kim, Park, and Choi}]{KorMedMCQA}
Sunjun Kweon, Byungjin Choi, Minkyu Kim, Rae~Woong Park, and Edward Choi. 2024.
\newblock \href {https://arxiv.org/abs/2403.01469} {Kormedmcqa: Multi-choice question answering benchmark for korean healthcare professional licensing examinations}.
\newblock \emph{arXiv}.

\bibitem[{Li et~al.(2021{\natexlab{a}})Li, Zhong, and Chen}]{li-etal-2021-mlec}
Jing Li, Shangping Zhong, and Kaizhi Chen. 2021{\natexlab{a}}.
\newblock \href {https://doi.org/10.18653/v1/2021.emnlp-main.698} {{MLEC-QA}: {A} {C}hinese {M}ulti-{C}hoice {B}iomedical {Q}uestion {A}nswering {D}ataset}.
\newblock In \emph{Proceedings of the 2021 Conference on Empirical Methods in Natural Language Processing}, pages 8862--8874, Online and Punta Cana, Dominican Republic. Association for Computational Linguistics.

\bibitem[{Li et~al.(2021{\natexlab{b}})Li, Zhong, and Chen}]{mlecqa}
Jing Li, Shangping Zhong, and Kaizhi Chen. 2021{\natexlab{b}}.
\newblock \href {https://doi.org/10.18653/v1/2021.emnlp-main.698} {{MLEC-QA}: {A} {C}hinese {M}ulti-{C}hoice {B}iomedical {Q}uestion {A}nswering {D}ataset}.
\newblock In \emph{Proceedings of the 2021 Conference on Empirical Methods in Natural Language Processing}, pages 8862--8874, Online and Punta Cana, Dominican Republic. Association for Computational Linguistics.

\bibitem[{Liu et~al.(2024)Liu, Okuhara, Chang, Shirabe, Nishiie, Okada, and Kiuchi}]{Liu2024}
M.~Liu, T.~Okuhara, X.~Chang, R.~Shirabe, Y.~Nishiie, H.~Okada, and T.~Kiuchi. 2024.
\newblock \href {https://doi.org/10.2196/60807} {Performance of chatgpt across different versions in medical licensing examinations worldwide: Systematic review and meta-analysis}.
\newblock \emph{Journal of Medical Internet Research}, 26:e60807.

\bibitem[{Pal et~al.(2022)Pal, Umapathi, and Sankarasubbu}]{medmcqa}
Ankit Pal, Logesh~Kumar Umapathi, and Malaikannan Sankarasubbu. 2022.
\newblock \href {https://arxiv.org/abs/2203.14371} {Medmcqa : A large-scale multi-subject multi-choice dataset for medical domain question answering}.

\bibitem[{Restrepo et~al.(2024{\natexlab{a}})Restrepo, Nakayama, Dychiao, Wu, McCoy, Artiaga, Cobanaj, Matos, Gallifant, Bitterman, Ferrer, Aphinyanaphongs, and Anthony~Celi}]{multiligual_restrepo}
David Restrepo, Luis~Filipe Nakayama, Robyn~Gayle Dychiao, Chenwei Wu, Liam~G. McCoy, Jose~Carlo Artiaga, Marisa Cobanaj, João Matos, Jack Gallifant, Danielle~S. Bitterman, Vincenz Ferrer, Yindalon Aphinyanaphongs, and Leo Anthony~Celi. 2024{\natexlab{a}}.
\newblock \href {https://doi.org/10.1109/ICHI61247.2024.00089} {Seeing beyond borders: Evaluating llms in multilingual ophthalmological question answering}.
\newblock In \emph{2024 IEEE 12th International Conference on Healthcare Informatics (ICHI)}, pages 565--566.

\bibitem[{Restrepo et~al.(2024{\natexlab{b}})Restrepo, Wu, Vásquez-Venegas, Matos, Gallifant, Celi, Bitterman, and Nakayama}]{restrepo_analyzing_2024}
David Restrepo, Chenwei Wu, Constanza Vásquez-Venegas, João Matos, Jack Gallifant, Leo~Anthony Celi, Danielle~S. Bitterman, and Luis~Filipe Nakayama. 2024{\natexlab{b}}.
\newblock \href {https://doi.org/10.1101/2024.06.18.24309113} {Analyzing {Diversity} in {Healthcare} {LLM} {Research}: {A} {Scientometric} {Perspective}}.

\bibitem[{Ryan et~al.(2024)Ryan, Held, and Yang}]{ryan_unintended_2024}
Michael~J. Ryan, William Held, and Diyi Yang. 2024.
\newblock \href {http://arxiv.org/abs/2402.15018} {Unintended {Impacts} of {LLM} {Alignment} on {Global} {Representation}}.
\newblock \emph{arXiv preprint}.
\newblock ArXiv:2402.15018 [cs].

\bibitem[{Saab et~al.(2024)Saab, Tu, Weng, Tanno, Stutz, Wulczyn, Zhang, Strother, Park, Vedadi, Chaves, Hu, Schaekermann, Kamath, Cheng, Barrett, Cheung, Mustafa, Palepu, McDuff, Hou, Golany, Liu, baptiste Alayrac, Houlsby, Tomasev, Freyberg, Lau, Kemp, Lai, Azizi, Kanada, Man, Kulkarni, Sun, Shakeri, He, Caine, Webson, Latysheva, Johnson, Mansfield, Lu, Rivlin, Anderson, Green, Wong, Krause, Shlens, Dominowska, Eslami, Chou, Cui, Vinyals, Kavukcuoglu, Manyika, Dean, Hassabis, Matias, Webster, Barral, Corrado, Semturs, Mahdavi, Gottweis, Karthikesalingam, and Natarajan}]{saab2024capabilitiesgeminimodelsmedicine}
Khaled Saab, Tao Tu, Wei-Hung Weng, Ryutaro Tanno, David Stutz, Ellery Wulczyn, Fan Zhang, Tim Strother, Chunjong Park, Elahe Vedadi, Juanma~Zambrano Chaves, Szu-Yeu Hu, Mike Schaekermann, Aishwarya Kamath, Yong Cheng, David G.~T. Barrett, Cathy Cheung, Basil Mustafa, Anil Palepu, Daniel McDuff, Le~Hou, Tomer Golany, Luyang Liu, Jean baptiste Alayrac, Neil Houlsby, Nenad Tomasev, Jan Freyberg, Charles Lau, Jonas Kemp, Jeremy Lai, Shekoofeh Azizi, Kimberly Kanada, SiWai Man, Kavita Kulkarni, Ruoxi Sun, Siamak Shakeri, Luheng He, Ben Caine, Albert Webson, Natasha Latysheva, Melvin Johnson, Philip Mansfield, Jian Lu, Ehud Rivlin, Jesper Anderson, Bradley Green, Renee Wong, Jonathan Krause, Jonathon Shlens, Ewa Dominowska, S.~M.~Ali Eslami, Katherine Chou, Claire Cui, Oriol Vinyals, Koray Kavukcuoglu, James Manyika, Jeff Dean, Demis Hassabis, Yossi Matias, Dale Webster, Joelle Barral, Greg Corrado, Christopher Semturs, S.~Sara Mahdavi, Juraj Gottweis, Alan Karthikesalingam, and Vivek Natarajan. 2024.
\newblock \href {https://arxiv.org/abs/2404.18416} {Capabilities of gemini models in medicine}.
\newblock \emph{arXiv}.

\bibitem[{Thirunavukarasu et~al.(2023)Thirunavukarasu, Ting, Elangovan, Gutierrez, Tan, and Ting}]{thirunavukarasu_large_2023}
Arun~James Thirunavukarasu, Darren Shu~Jeng Ting, Kabilan Elangovan, Laura Gutierrez, Ting~Fang Tan, and Daniel Shu~Wei Ting. 2023.
\newblock \href {https://doi.org/10.1038/s41591-023-02448-8} {Large language models in medicine}.
\newblock \emph{Nature Medicine}, 29(8):1930--1940.

\bibitem[{Vilares and G{\'o}mez-Rodr{\'\i}guez(2019{\natexlab{a}})}]{vilares-gomez-rodriguez-2019-head}
David Vilares and Carlos G{\'o}mez-Rodr{\'\i}guez. 2019{\natexlab{a}}.
\newblock \href {https://doi.org/10.18653/v1/P19-1092} {{HEAD}-{QA}: A healthcare dataset for complex reasoning}.
\newblock In \emph{Proceedings of the 57th Annual Meeting of the Association for Computational Linguistics}, pages 960--966, Florence, Italy. Association for Computational Linguistics.

\bibitem[{Vilares and G{\'o}mez-Rodr{\'\i}guez(2019{\natexlab{b}})}]{headqa}
David Vilares and Carlos G{\'o}mez-Rodr{\'\i}guez. 2019{\natexlab{b}}.
\newblock \href {https://doi.org/10.18653/v1/P19-1092} {{HEAD}-{QA}: A healthcare dataset for complex reasoning}.
\newblock In \emph{Proceedings of the 57th Annual Meeting of the Association for Computational Linguistics}, pages 960--966, Florence, Italy. Association for Computational Linguistics.

\bibitem[{Wang et~al.(2023)}]{wang2023culturalvqa}
Yiyi Wang et~al. 2023.
\newblock Culturalvqa: A new frontier in vision and language understanding.
\newblock In \emph{Proceedings of CVPR}.

\bibitem[{Weidinger et~al.(2021)}]{weidinger2021ethical}
Laura Weidinger et~al. 2021.
\newblock Ethical and social risks of harm from language models.
\newblock \emph{arXiv preprint arXiv:2112.04359}.

\bibitem[{Wiggers(2024)}]{wiggers_hugging_2024}
Kyle Wiggers. 2024.
\newblock \href {https://techcrunch.com/2024/04/18/hugging-face-releases-a-benchmark-for-testing-generative-ai-on-health-tasks/} {Hugging {Face} releases a benchmark for testing generative {AI} on health tasks}.

\bibitem[{Wu et~al.(2023)Wu, Lei, Zheng, Zhao, Lin, Zhang, Zhou, Zhao, Zhang, Wang, and Xie}]{wu2023gpt4visionservemedicalapplications}
Chaoyi Wu, Jiayu Lei, Qiaoyu Zheng, Weike Zhao, Weixiong Lin, Xiaoman Zhang, Xiao Zhou, Ziheng Zhao, Ya~Zhang, Yanfeng Wang, and Weidi Xie. 2023.
\newblock \href {https://arxiv.org/abs/2310.09909} {Can gpt-4v(ision) serve medical applications? case studies on gpt-4v for multimodal medical diagnosis}.
\newblock \emph{Preprint}, arXiv:2310.09909.

\bibitem[{Yan et~al.(2023)Yan, Zhang, Zhou, He, Li, and Sun}]{yan2023multimodalchatgptmedicalapplications}
Zhiling Yan, Kai Zhang, Rong Zhou, Lifang He, Xiang Li, and Lichao Sun. 2023.
\newblock \href {https://arxiv.org/abs/2310.19061} {Multimodal chatgpt for medical applications: an experimental study of gpt-4v}.
\newblock \emph{Preprint}, arXiv:2310.19061.

\bibitem[{Yu et~al.(2024)Yu, Fan, Li, Zhou, Ma, Xian, Hua, He, Jin, Zhang, Gandhi, and Ma}]{yu2024large}
Huizi Yu, Lizhou Fan, Lingyao Li, Jiayan Zhou, Zihui Ma, Lu~Xian, Wenyue Hua, Sijia He, Mingyu Jin, Yongfeng Zhang, Ashvin Gandhi, and Xin Ma. 2024.
\newblock \href {https://arxiv.org/abs/2403.16303} {Large language models in biomedical and health informatics: A bibliometric review}.
\newblock \emph{Preprint}, arXiv:2403.16303.

\bibitem[{Zack et~al.(2024)Zack, Lehman, Suzgun, Rodriguez, Celi, Gichoya, Jurafsky, Szolovits, Bates, Abdulnour, Butte, and Alsentzer}]{zack_assessing_2024}
Travis Zack, Eric Lehman, Mirac Suzgun, Jorge~A Rodriguez, Leo~Anthony Celi, Judy Gichoya, Dan Jurafsky, Peter Szolovits, David~W Bates, Raja-Elie~E Abdulnour, Atul~J Butte, and Emily Alsentzer. 2024.
\newblock \href {https://doi.org/10.1016/S2589-7500(23)00225-X} {Assessing the potential of {GPT}-4 to perpetuate racial and gender biases in health care: a model evaluation study}.
\newblock \emph{The Lancet Digital Health}, 6(1):e12--e22.

\bibitem[{Zhang et~al.(2024{\natexlab{a}})Zhang, Klyman, Mai, Levine, Zhang, Bommasani, and Liang}]{zhang_language_2024}
Andy~K. Zhang, Kevin Klyman, Yifan Mai, Yoav Levine, Yian Zhang, Rishi Bommasani, and Percy Liang. 2024{\natexlab{a}}.
\newblock \href {https://doi.org/10.48550/arXiv.2410.08385} {Language model developers should report train-test overlap}.
\newblock \emph{arXiv preprint}.
\newblock ArXiv:2410.08385 [cs].

\bibitem[{Zhang et~al.(2024{\natexlab{b}})Zhang, Da, Lee, Robinson, Wu, Song, Zhao, Raja, Slack, Lyu, Hendryx, Kaplan, Lunati, and Yue}]{zhang_careful_2024}
Hugh Zhang, Jeff Da, Dean Lee, Vaughn Robinson, Catherine Wu, Will Song, Tiffany Zhao, Pranav Raja, Dylan Slack, Qin Lyu, Sean Hendryx, Russell Kaplan, Michele Lunati, and Summer Yue. 2024{\natexlab{b}}.
\newblock \href {https://doi.org/10.48550/arXiv.2405.00332} {A {Careful} {Examination} of {Large} {Language} {Model} {Performance} on {Grade} {School} {Arithmetic}}.
\newblock \emph{arXiv preprint}.
\newblock ArXiv:2405.00332 [cs].

\bibitem[{Zhang et~al.(2023{\natexlab{a}})}]{zhang2023examsv}
Jieyi Zhang et~al. 2023{\natexlab{a}}.
\newblock Exams-v: A multi-discipline multi-lingual multi-modal exam benchmark for evaluating vision language models.
\newblock \emph{arXiv preprint arXiv:2308.03463}.

\bibitem[{Zhang et~al.(2023{\natexlab{b}})Zhang, Yang, Zhang et~al.}]{zhang2023mmmu}
Xiang Zhang, Junyang Yang, Jianwei Zhang, et~al. 2023{\natexlab{b}}.
\newblock Mmmu: A massive multi-discipline multimodal understanding and reasoning benchmark for expert agi.
\newblock In \emph{Proceedings of NeurIPS}.

\bibitem[{Üstün et~al.(2024)Üstün, Aryabumi, Yong, Ko, D'souza, Onilude, Bhandari, Singh, Ooi, Kayid, Vargus, Blunsom, Longpre, Muennighoff, Fadaee, Kreutzer, and Hooker}]{ustun2024ayamodelinstructionfinetuned}
Ahmet Üstün, Viraat Aryabumi, Zheng-Xin Yong, Wei-Yin Ko, Daniel D'souza, Gbemileke Onilude, Neel Bhandari, Shivalika Singh, Hui-Lee Ooi, Amr Kayid, Freddie Vargus, Phil Blunsom, Shayne Longpre, Niklas Muennighoff, Marzieh Fadaee, Julia Kreutzer, and Sara Hooker. 2024.
\newblock \href {https://arxiv.org/abs/2402.07827} {Aya model: An instruction finetuned open-access multilingual language model}.
\newblock \emph{Preprint}, arXiv:2402.07827.

\end{thebibliography}

% Appendix
\onecolumn
\appendix
\section{Appendix}
\subsection{Existing publicly-available medical examination QA dataset per country}
\label{sec:medqas}
\begin{table}[ht]
\caption{Summary of existing open-source Medical QA Datasets by Country.}
\small
\begin{tabular}{|m{1.2cm}|m{2.3cm}|m{1cm}|m{2cm}|m{2cm}|m{2.5cm}|m{1.5cm}|}
\hline
\textbf{Country} & \textbf{Dataset} & \textbf{\#QA} & \textbf{Language(s)} & \textbf{Modalities} & \textbf{Source} &  \textbf{Years} \\ \hline
China & MedQA \cite{medqa} & 34,251 & Simplified Chinese & Text & MCMLE, Mainland China Medical Licensing Examination & Not Clear \\ \hline
China & MLEC-QA \cite{mlecqa} & 136,236 & Simplified Chinese & Text, Images & National Medical Licensing Examination (NMLEC) & Not Clear \\ \hline
India & MedMCQA \cite{medmcqa} & 193,155 & English & Text & AIIMS PG, NEET PG & 1991-2022 \\ \hline
Spain & Head-QA \cite{headqa} & 6,765 & Spanish and English & Text, Images & Ministerio de Sanidad, Consumo y Bienestar Social & 2013–2017  \\ \hline
Republic of Korea & KorMedMCQA \cite{KorMedMCQA} & 5,345 & Korean and English & Text & Korea Health Personnel Licensing Examination Institute & 2012–2023  \\ \hline
Sweden & MedQA-SWE \cite{swemedqa} & 3,180 & Swedish & Text & National Board of Health and Welfare, Umeå University & 2016–2023 \\ \hline
Taiwan & MedQA \cite{medqa} & 14,123 & Traditional Chinese & Text & TWMLE, Taiwan Medical Licensing Examination & Not Clear  \\ \hline
United States & MedQA \cite{medqa} & 12,723 & English & Text & USMLE, United States Medical Licensing Examination & Not Clear \\ \hline
\end{tabular}
\label{tab:medqa_datasets}
\end{table}

\newpage
\subsection{Details on collected data per country}
\label{sec:data_details}

\subsubsection{Brazil} The examination data were collected from the "Revalida" examinations, which are publicly available on the Brazilian government's official website. The "Revalida" exam, administered by the National Institute of Educational Research and Studies (INEP), supports the process of diploma revalidation for doctors who graduated abroad and wish to practice in Brazil. The exams consist of two sections: 100 multiple-choice questions (20 in each of the following areas: Internal Medicine, Surgery, Pediatrics, Preventive Medicine, and Gynecology and Obstetrics) and open-ended questions. For this work, only the multiple-choice section was included. Data from the years 2011–2016 and 2020–2024 were used, encompassing all publicly available years at the time of this study \cite{brazil_data}.

\subsubsection{Israel} 
The Israeli subset consists of questions from seven medical specialties: Internal Medicine, Clinical Microbiology, Neurology, Oncology, Ophthalmology, Urology, and Public Health. These questions are drawn from Phase A of the two-phase examination process that residents in Israel must complete during their training. Phase A is a written exam held annually, comprising approximately 150 questions. We included questions from tests administered between 2020 and 2023, with full versions of these exams publicly available on the Israel Medical Association's website \cite{ima_english, hebrew_data}.

The questions are categorized into three main types: preclinical cases, clinical cases, and questions based on scientific articles. Preclinical cases focus on foundational scientific knowledge, while clinical cases present patient background information followed by clinical questions related to the patient's medical conditions. Questions derived from scientific articles involve analysis of graphs, figures, and study results, which are particularly prevalent in the public health exam. Some questions also include visual aids, such as diagnostic images, laboratory slides (e.g., blood smear slides in Clinical Microbiology), and other data specific to patients' clinical presentations.

\subsubsection{Japan} Japanese questions were sourced from past examinations that were published on the website of the Ministry of Health, Labour and Welfare. We included the 116th, 117th, and 118th National Medical Licensing Examination \cite{jp116, jp117, jp118}, which corresponded to 2022-2024.

\subsubsection{Spain} The examination data were sourced from the annual exams organized by the Ministerio de Sanidad, Consumo y Bienestar Social (Spanish Ministry of Health, Consumer Affairs, and Social Welfare) \cite{spain_data}. These exams are part of the competitive selection process for specialized medical positions in Spain's public healthcare system. Eligibility for participation requires candidates to possess a bachelor's degree in medicine (6 years of study) and typically prepare for a year or more, given the limited number of vacancies. The exams play a critical role in ranking candidates, who are able to select their specialization and hospital placement only based on their exam performance. For this study, only data from the years 2019–2023 were used to avoid overlap with the existing Head-QA dataset \cite{headqa}.

\newpage
\subsection{Detailed Data Statistics}
\label{sec:data_stats}

\begin{table*}[ht]
\centering
\caption{\texttt{WorldMedQA's} data across countries and languages. In the curated dataset, each QA was associated with at least one image. Some images were present in more than one question. In the final subset for evaluation (rightmost column), each question had a single image and the correct option associated with it, resulting in fewer samples. The number of answer options per question (fourth column) refers to the original number of choices in the multiple-choice format (e.g., A-D for four options or A-E for five options). However, all questions after preprocessing results in 4 options only. In cases where this varies, such as in Brazil, the value represents a weighted average across questions. The total number of QAs and images does not immediately add up due to some questions sharing images or having multiple associated options.}
\label{tab:data}
\begin{tabular}{|l|l|c|c|c|c|c|}
\hline
\textbf{Country} & \textbf{Language} & \textbf{Years} & \textbf{Option/QA} & \textbf{QAs, n (\%)} & \textbf{Images, n (\%)} & \textbf{Final, n (\%)} \\ \hline
Brazil  & Portuguese & 2011-2024 & 4.27 & 93 (12.8\%)  & 94 (11.1\%)  & 89 (15.7\%) \\ \hline
Israel  & Hebrew     & 2020-2023 & 4.00  & 200 (27.6\%) & 184 (21.6\%) & 186 (32.7\%) \\ \hline
Japan   & Japanese   & 2022-2024 & 5.00  & 306 (42.1\%) & 445 (52.4\%) & 168 (29.6\%) \\ \hline
Spain   & Spanish    & 2019-2023 & 4.00  & 127 (17.5\%) & 127 (14.9\%) & 125 (22.0\%) \\ \hline
\textbf{Total}   & 4 Languages & 2011-2024 & 4.00    & 726 (100\%) & 850 (100\%) & 568 (100\%) \\ \hline
\end{tabular}
\end{table*}

\newpage
\subsection{Example QA from the Brazilian dataset}
\label{sec:example}

\begin{tcolorbox}[colframe=black, colback=white, title=Box 1. Example multimodal QA from the Brazilian subset]
    \begin{multicols}{3}
    
    % Left panel (Portuguese)
    \columnbreak
    \noindent
    \textbf{Original (Portuguese)}\\
    Um paciente do sexo masculino, 55 anos de idade, tabagista 60 maços/ano, com tosse crônica há mais de 10 anos, relata que há cerca de três meses observou a presença de sangue na secreção eliminada com a tosse. Refere ainda perda de cerca de 15\% do peso habitual nesse mesmo período, anorexia, adinamia e sudorese noturna. A radiografia de tórax realizada por ocasião da consulta é mostrada abaixo. Qual a hipótese diagnóstica mais provável nesse caso?\\
    
    A) Aspergilose pulmonar.\\
    \textbf{B) Carcinoma pulmonar.}\\
    C) Tuberculose cavitária.\\
    D) Bronquiectasia com infecção.\\
    E) Doença pulmonar obstrutiva crônica.\\
    
    % Middle panel (Image)
    \columnbreak
    \noindent
    \textbf{Image}\\
    \includegraphics[width=0.9\columnwidth]{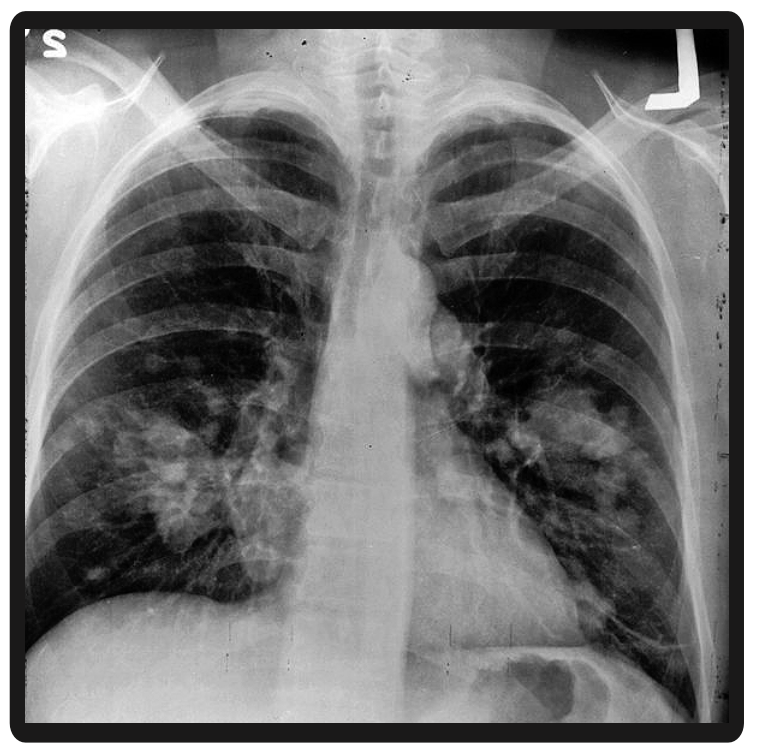} 
    
    % Right panel (English Translation)
    \columnbreak
    \noindent
    \textbf{Translation (English)}\\
    A 55-year-old male patient, with a smoking history of 60 pack-years, has had a chronic cough for over 10 years. He reports that about three months ago, he noticed the presence of blood in the sputum. He also mentions a weight loss of about 15\% of his usual weight during the same period, anorexia, weakness, and night sweats. The chest X-ray taken at the time of the consultation is shown below. What is the most likely diagnostic hypothesis in this case?\\
    
    A) Pulmonary aspergillosis.\\
    \textbf{B) Lung carcinoma.}\\
    C) Cavitary tuberculosis.\\
    D) Bronchiectasis with infection.\\
    E) Chronic obstructive pulmonary disease.\\
    
    \end{multicols}
\end{tcolorbox}

\newpage
\subsection{Model output consistency across countries and test setting}
\label{sec:kappas}
\begin{table}[h]
\centering
\caption{Cohen's Kappa reflecting agreement between languages for the same models, countries, and testing setting. Values in \textbf{bold} highlight the model with highest kappa per country and testing mode. The two studied settings were text-only (T. only) and text and image (T. \& I.).}
\label{tab:agreement_images}
\begin{tabular}{|l|c|c|c|c|c|c|c|c|}
\hline
\textbf{Country} & \multicolumn{2}{c|}{\textbf{Brazil}} & \multicolumn{2}{c|}{\textbf{Israel}} & \multicolumn{2}{c|}{\textbf{Japan}} & \multicolumn{2}{c|}{\textbf{Spain}} \\ \hline
\textbf{Model $\downarrow$  Setting $\rightarrow$} & \textbf{T. only} & \textbf{T. \& I.} & \textbf{T. only} & \textbf{T. \& I.} & \textbf{T. only} & \textbf{T. \& I.} & \textbf{T. only} & \textbf{T. \& I.} \\ \hline
\textit{GPT4o} & \textbf{0.684} & \textbf{0.743} & \textbf{0.619} & 0.654 & \textbf{0.683} & \textbf{0.618} & \textbf{0.840} & \textbf{0.829} \\ \hline
\textit{GPT4o-MINI} & 0.642 & 0.655 & 0.458 & 0.603 & 0.525 & 0.554 & 0.715 & 0.809 \\ \hline
\textit{GeminiFlash1-5} & 0.612 & 0.536 & 0.533 & \textbf{0.655} & 0.591 & 0.521 & 0.594 & 0.767 \\ \hline
\textit{GeminiPro1-5} & 0.389 & 0.469 & 0.184 & 0.416 & 0.351 & 0.490 & 0.163 & 0.693 \\ \hline
\textit{Yi-VL-34B} & 0.020 & 0.438 & 0.111 & 0.309 & 0.033 & 0.393 & 0.030 & 0.507 \\ \hline
\textit{Yi-VL-6B} & 0.427 & 0.320 & 0.150 & 0.204 & 0.240 & 0.348 & 0.269 & 0.251 \\ \hline
\textit{llava-next-llama3} & 0.429 & 0.441 & 0.401 & 0.380 & 0.269 & 0.264 & 0.435 & 0.433 \\ \hline
\textit{llava-next-mistral-7b} & 0.498 & 0.310 & 0.148 & 0.243 & 0.153 & 0.234 & 0.466 & 0.348 \\ \hline
\textit{llava-next-vicuna-7b} & 0.385 & 0.491 & 0.167 & 0.281 & 0.091 & 0.185 & 0.310 & 0.279 \\ \hline
\textit{llava-next-yi-34b} & 0.592 & 0.635 & 0.208 & 0.223 & 0.393 & 0.373 & 0.594 & 0.488 \\ \hline
\end{tabular}
\end{table}

\newpage
\subsection{Detailed performance with and without images}
\label{sec:detail_performance}
\begin{table}[h]
\centering
\caption{Accuracy comparison across countries and \textbf{original} languages (Portuguese, Hebrew, Japanese, and Spanish) for each model. The two studied settings were text-only (T. only) and text and image (T. \& I.). Each cell represents the performance of each model in its native language dataset, highlighting how the presence or absence of images affects accuracy.}
\label{tab:agreement_images}
\begin{tabular}{|l|c|c|c|c|c|c|c|c|}
\hline
\textbf{Country} & \multicolumn{2}{c|}{\textbf{Brazil}} & \multicolumn{2}{c|}{\textbf{Israel}} & \multicolumn{2}{c|}{\textbf{Japan}} & \multicolumn{2}{c|}{\textbf{Spain}} \\ \hline
\textbf{Model $\downarrow$  Setting $\rightarrow$} & \textbf{T. only} & \textbf{T. \& I.} & \textbf{T. only} & \textbf{T. \& I.} & \textbf{T. only} & \textbf{T. \& I.} & \textbf{T. only} & \textbf{T. \& I.} \\ \hline
\textit{GPT4o} & 0.764 & 0.753 & 0.584 & 0.578 & 0.857 & 0.881 & 0.704 & 0.712 \\ \hline
\textit{GPT4o-MINI} & 0.562 & 0.584 & 0.389 & 0.373 & 0.762 & 0.732 & 0.632 & 0.640 \\ \hline
\textit{GeminiFlash1-5} & 0.494 & 0.640 & 0.281 & 0.395 & 0.625 & 0.667 & 0.456 & 0.640 \\ \hline
\textit{GeminiPro1-5} & 0.427 & 0.607 & 0.135 & 0.368 & 0.601 & 0.726 & 0.312 & 0.584 \\ \hline
\textit{Yi-VL-34B} & 0.034 & 0.438 & 0.011 & 0.270 & 0.077 & 0.530 & 0.024 & 0.416 \\ \hline
\textit{Yi-VL-6B} & 0.348 & 0.348 & 0.238 & 0.249 & 0.446 & 0.482 & 0.328 & 0.336 \\ \hline
\textit{llava-next-llama3} & 0.393 & 0.382 & 0.330 & 0.297 & 0.458 & 0.470 & 0.368 & 0.416 \\ \hline
\textit{llava-next-mistral-7b} & 0.371 & 0.404 & 0.238 & 0.265 & 0.369 & 0.381 & 0.376 & 0.336 \\ \hline
\textit{llava-next-vicuna-7b} & 0.281 & 0.292 & 0.276 & 0.308 & 0.256 & 0.298 & 0.304 & 0.328 \\ \hline
\textit{llava-next-yi-34b} & 0.438 & 0.483 & 0.238 & 0.281 & 0.577 & 0.518 & 0.504 & 0.488 \\ \hline
\end{tabular}
\end{table}
\begin{table}[h]
\centering
\caption{Accuracy comparison across countries and \textbf{English-translated} datasets for each model. The two studied settings were text-only (T. only) and text and image (T. \& I.). Each cell represents the performance of each model after translation, highlighting how the presence or absence of images affects accuracy.}
\label{tab:agreement_images_english}
\begin{tabular}{|l|c|c|c|c|c|c|c|c|}
\hline
\textbf{Country} & \multicolumn{2}{c|}{\textbf{Brazil}} & \multicolumn{2}{c|}{\textbf{Israel}} & \multicolumn{2}{c|}{\textbf{Japan}} & \multicolumn{2}{c|}{\textbf{Spain}} \\ \hline
\textbf{Model $\downarrow$  Setting $\rightarrow$} & \textbf{T. only} & \textbf{T. \& I.} & \textbf{T. only} & \textbf{T. \& I.} & \textbf{T. only} & \textbf{T. \& I.} & \textbf{T. only} & \textbf{T. \& I.} \\ \hline
\textit{GPT4o} & 0.685 & 0.674 & 0.589 & 0.632 & 0.690 & 0.720 & 0.720 & 0.728 \\ \hline
\textit{GPT4o-MINI} & 0.517 & 0.584 & 0.422 & 0.438 & 0.595 & 0.613 & 0.592 & 0.632 \\ \hline
\textit{GeminiFlash1-5} & 0.494 & 0.539 & 0.286 & 0.427 & 0.571 & 0.601 & 0.544 & 0.632 \\ \hline
\textit{GeminiPro1-5} & 0.382 & 0.652 & 0.281 & 0.492 & 0.440 & 0.613 & 0.248 & 0.632 \\ \hline
\textit{Yi-VL-34B} & 0.034 & 0.472 & 0.022 & 0.389 & 0.065 & 0.536 & 0.024 & 0.544 \\ \hline
\textit{Yi-VL-6B} & 0.326 & 0.348 & 0.335 & 0.362 & 0.393 & 0.417 & 0.392 & 0.448 \\ \hline
\textit{llava-next-llama3} & 0.494 & 0.483 & 0.314 & 0.297 & 0.435 & 0.482 & 0.568 & 0.552 \\ \hline
\textit{llava-next-mistral-7b} & 0.438 & 0.416 & 0.319 & 0.319 & 0.500 & 0.452 & 0.496 & 0.512 \\ \hline
\textit{llava-next-vicuna-7b} & 0.371 & 0.337 & 0.303 & 0.314 & 0.393 & 0.399 & 0.472 & 0.448 \\ \hline
\textit{llava-next-yi-34b} & 0.528 & 0.539 & 0.459 & 0.432 & 0.577 & 0.577 & 0.600 & 0.592 \\ \hline
\end{tabular}
\end{table}

\end{document}